\documentclass[twocolumn]{article}
\usepackage[margin={2.1cm,2.1cm}]{geometry}

\newcommand{\sfigwhole}[0]{17.0cm}
\newcommand{\sfigbig}[0]{5.2cm}
\newcommand{\sfigmedium}[0]{4.0cm}

\newcommand{\sfigsmallsmall}[0]{3.7cm}

\newcommand{\sfigspecialI}[0]{4.5cm}
\newcommand{\sfigspecialII}[0]{6.0cm}
\newcommand{\sfigspecialIII}[0]{5.5cm}
\newcommand{\introvspace}[0]{0pt}

\newcommand{\class}{{c}}

\newcommand{\latentVector}{{Z}}

\newcommand{\conditionedLatentVector}{{\latentVector_{\class}}}

\newcommand{\generator}{{G}}
\newcommand{\discriminator}{{D}}
\newcommand{\image}{{X}}
\newcommand{\conditionedImage}{{\image_{\class}}}

\newcommand{\conditionedFakeImage}{{{\image_{\class}}}}

\newcommand{\discriminatorEncode}{{\discriminator_{e}}}
\newcommand{\discriminatorDiscriminate}{{\discriminator_{d}}}

\newcommand{\decoder}{{\Delta}}
\newcommand{\encoder}{{E}}

\newcommand{\meanSymbol}{\mu}
\newcommand{\covarianceSymbol}{\Sigma}
\newcommand{\conditionalLatentDistribution}{\mathcal{N}_{\class}}

\newcommand{\minorityDigit}{{0}}
\newcommand{\digitDropOut}{{97.5\%}}

\usepackage{flushend}       
\usepackage{dblfloatfix}    

\newcommand{\flowers}{{\textit{Flowers}}}
\newcommand{\cifar}{{\textit{CIFAR-10}}}
\newcommand{\gts}{{\textit{GTSRB}}}
\newcommand{\mnist}{{\textit{MNIST}}}

\newcommand{\acgan}{{\textit{ACGAN}}}
\newcommand{\gan}{{\textit{GAN}}}
\newcommand{\bagan}{{\textit{BAGAN}}}

\newcommand{\ssimname}{{SSIM}}
\newcommand{\ssim}{{\ssimname}}

\newcommand{\resnet}{{\textit{ResNet-18}}}

\usepackage{multirow}
\usepackage{enumerate}

\usepackage{microtype}
\usepackage{graphicx}
\usepackage{subfigure}
\usepackage{booktabs} 
\usepackage{color}

\usepackage{hyperref}

\title{BAGAN: Data Augmentation with Balancing GAN}
\author{Giovanni Mariani, Florian Scheidegger, Roxana Istrate, Costas Bekas, and Cristiano Malossi\\
IBM Research -- Zurich, Switzerland}
\date{}

\begin{document}
\maketitle
\let\thefootnote\relax\footnotetext{IBM, the IBM~logo, and ibm.com are trademarks or registered trademarks of International Business Machines Corporation in the United States, other countries, or both. Other product and service names might be trademarks of IBM or other companies. Submitted. Copyright 2018 by the author(s).}

\begin{abstract}
Image classification datasets are often imbalanced, characteristic that negatively affects the accuracy of deep-learning classifiers. In this work we propose balancing GAN (BAGAN) as an augmentation tool to restore balance in imbalanced datasets. This is challenging because the few minority-class images may not be enough to train a GAN. We overcome this issue by including during the adversarial training all available images of majority and minority classes. The generative model learns useful features from majority classes and uses these to generate images for minority classes. We apply class conditioning in the latent space to drive the generation process towards a target class. The generator in the GAN is initialized with the encoder module of an autoencoder that enables us to learn an accurate class-conditioning in the latent space. We compare the proposed methodology with state-of-the-art GANs and demonstrate that BAGAN generates images of superior quality when trained with an imbalanced dataset.


\end{abstract}
\vspace{\introvspace}
\section{Introduction}
\label{sec:intro}

The accuracy of image classification techniques
can significantly deteriorate when the training dataset is imbalanced,
i.e. when available data is not uniformly distributed between the different classes.
Imbalanced datasets are common 
and a traditional approach to mitigate this problem
is to augment the dataset by introducing additional minority-class images derived
by applying simple geometric transformations to original images,
e.g. rotations or mirroring.
This augmentation approach may disrupt orientation-related features
when these are relevant.
In this work we propose a
balancing generative adversarial network (BAGAN) as an
augmentation tool to restore the dataset balance by generating new minority-class images.
Since these images are scarce in the initial dataset, it is challenging
to train a GAN for generating new ones.
To overcome this issue, the proposed methodology
includes in the adversarial training all data from minority and majority classes at
once.
This enables BAGAN to learn underlying features of the specific classification
problem starting from all images and then to apply these features for the generation
of new minority-class images. For example, let us consider the
classification of road traffic signs \cite{gts}. 
All warning signs share the same external triangular shape.
Once BAGAN learns to draw this
shape from one of the signs,
we can apply it for drawing any other one.
Since BAGAN learns features starting from all classes whereas the goal is to generate
images for the minority classes, a mechanism to drive the generation process toward
a desired class is needed. To this end,
in this work we apply class conditioning on the latent space \cite{odena2017,nguyen2016}.
We initialize the discriminator and generator in the GAN with an autoencoder. Then, we leverage this autoencoder to learn
class conditioning in the latent space, i.e. to learn how the input of the generative model should look like for different classes.
Additionally, this initialization enables us to start the adversarial training from a more stable point and helps mitigating convergence problems arising with traditional GANs \cite{srivastava2017,roth2017,kodali2017,lucic2017}. The main contributions of this work are:
\begin{itemize}
\item An overall approach to train GANs with an imbalanced dataset while specifically aiming to generate minority-class images.
\item An autoencoder-based initialization strategy that enables us to \textit{a)} start training the GAN from a good initial solution, and \textit{b)} learn how to encode different classes in the latent space of the generator.
\item An empirical evaluation of the proposed BAGAN methodology against the state of the art.
\end{itemize}


Experimental results empirically demonstrate that the proposed BAGAN methodology
outperforms state-of-the-art GAN approaches in terms of variety and quality
of the generated images when the training dataset is imbalanced. In turn, this leads
to a higher accuracy of final classifiers trained on the augmented dataset.

\section{Background}
\label{sec:back}

In recent years generative adversarial neural networks (GANs)
\cite{goodfellow2014,radford2015,lucic2017} have been proposed as a tool
to artificially generate realistic images.
The underlying idea is to train a generative network
in adversarial mode against a discriminator network.

A well known problem of generative adversarial models is that while they learn to fool the discriminator they may end up drawing one or few foolish examples. This problem is known
as mode collapse \cite{goodfellow2014,srivastava2017,roth2017,kodali2017}.
In this work, our aim is to augment an imbalanced image classification dataset
to restore its balance.
It is of paramount importance that the augmented dataset
is variable enough and does not include a continuously repeating
example, thus we need to avoid mode collapse.
To this end, different approaches have been proposed. Possible solutions are:
explicitly promoting image diversity in the generator loss \cite{lin2017,srivastava2017},
letting the generator
predict future changes of the discriminator and adapt against these
\cite{metz2017},
let the discriminator distinguish the different classes \cite{odena2017,salimans2016},
applying specific regularization techniques \cite{roth2017,kodali2017},
and coupling GANs with autoencoders \cite{srivastava2017,nguyen2016a,dumoulin2016,donahue2016}.

In this work we apply the latter approach and couple GAN and autoencoding techniques.
The cited approaches include additional modules in the GAN to embed an autoencoder all along the training. 
In the proposed BAGAN methodology
we apply a more pragmatic approach and use an autoencoder to initialize
the GAN modules close to a good solution and far from mode collapse.
Since our goal is to generate images specifically for the minority classes,
we train a generator that is controllable in terms of the
image class it draws, similarly to the state-of-the-art ACGAN methodology \cite{odena2017}.
Nonetheless, ACGAN is not specifically meant
for imbalanced datasets and turns to be flawed when targeting the generation of minority-class images.


\section{Motivating Example}
\label{sec:mot}

State-of-the-art GANs are not suitable to deal with
imbalanced datasets \cite{perez2017} and, to the best of our knowledge,
the proposed BAGAN methodology is the first one to specifically address this topic.
Before going through details of the proposed approach, let us demonstrate with a simple
example why it is difficult to apply existing GAN techniques
to tackle the problem at hand. 
Let us consider the classification of handwritten digits,
starting from an imbalanced version of the MNIST dataset \cite{mnist} where we remove
$\digitDropOut$ of the available zeros from the training set. 

A trivial idea would be to use a traditional GAN \cite{goodfellow2014,radford2015,karras2018},
train it by using all the available data, generate many random samples,
find the $\minorityDigit$ instances and use these for augmenting the dataset.
This approach cannot be applied in general:
if the generator $\generator$ in the GAN
is trained to fool the discriminator $\discriminator$
by generating realistic images, it will better focus on the
generation of majority classes to optimize its loss function while
collapsing away the modes related to the minority class.
On the other hand, training a GAN by using only the minority-class images is not really an option because minority-class images are scarce.
In this example, after removing $\digitDropOut$ of the zeros,
we are left with about 150 minority-class images.
In general, it is difficult to train a GAN starting from a very little dataset,
the GAN must have many examples to learn from \cite{gurumurthy2017}.

\begin{figure}[b!]
\centering
\subfigure[ACGAN discriminator.]{
            \label{fig:motACGAN}
            \resizebox{\sfigsmallsmall}{!}{\includegraphics{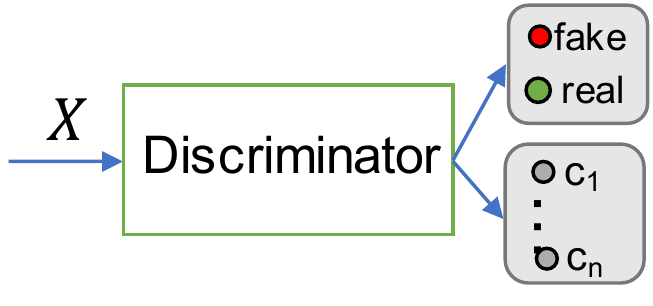}}
            }
\subfigure[BAGAN discriminator.]{
          \label{fig:motBAGAN}
            \resizebox{\sfigsmallsmall}{!}{\includegraphics{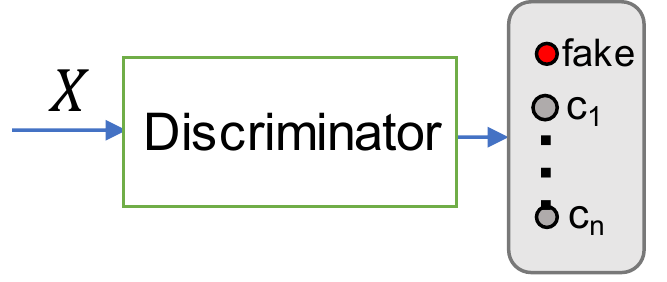}}
            }
\caption{Discriminator architectures for ACGAN and BAGAN.
}
\label{fig:motivatingDiscriminators}
\end{figure}

\begin{figure}[b!]
\centering
\subfigure[ACGAN.]{
          \label{fig:motResACGAN}
            \resizebox{\sfigbig}{!}{\includegraphics{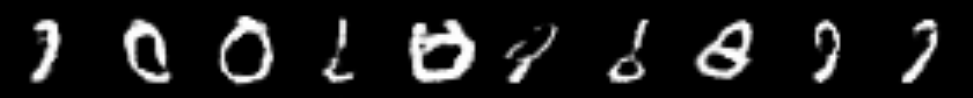}}
            }
\subfigure[Proposed BAGAN.]{
            \label{fig:motResBAGAN}
            \resizebox{\sfigbig}{!}{\includegraphics{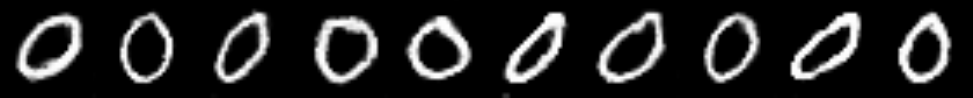}}
            }
\caption{Ten zero-digit images generated with ACGAN and the proposed BAGAN when trained with
an imbalanced version of MNIST where $\digitDropOut$
of the zeros were dropped.
}
\label{fig:motivatingResultsRef}
\end{figure}

\begin{figure*}[b!]
\centering
\subfigure[Autoencoder training.]{
            \label{fig:metAutoencoder}
            \resizebox{\sfigspecialI}{!}{\includegraphics{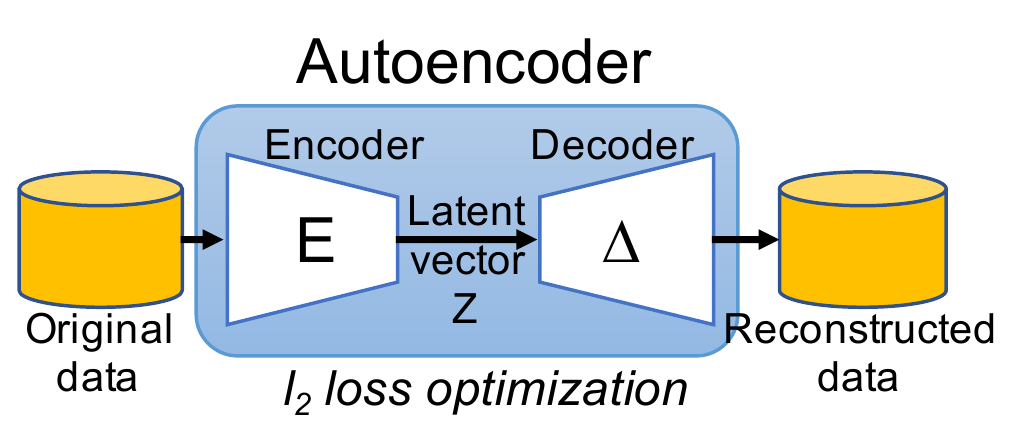}}
            }
\subfigure[GAN initialization.]{
          \label{fig:metInit}
            \resizebox{\sfigspecialII}{!}{\includegraphics{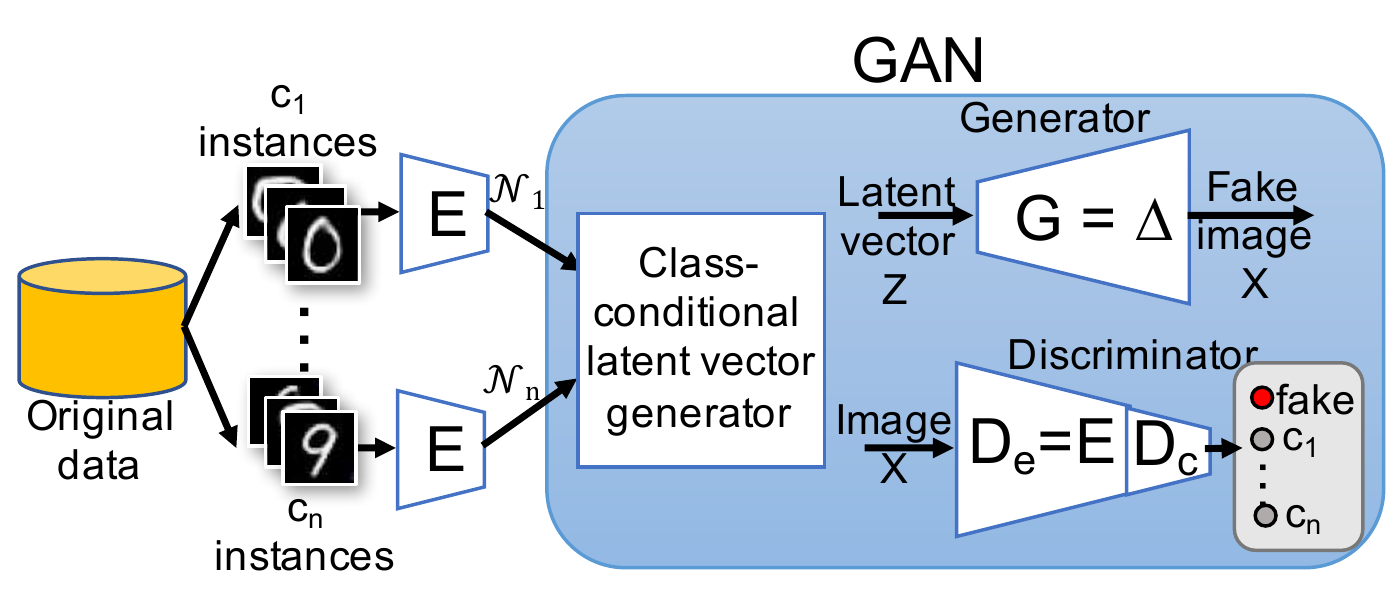}}
            }
\subfigure[GAN training.]{
            \label{fig:metGAN}
            \resizebox{\sfigspecialIII}{!}{\includegraphics{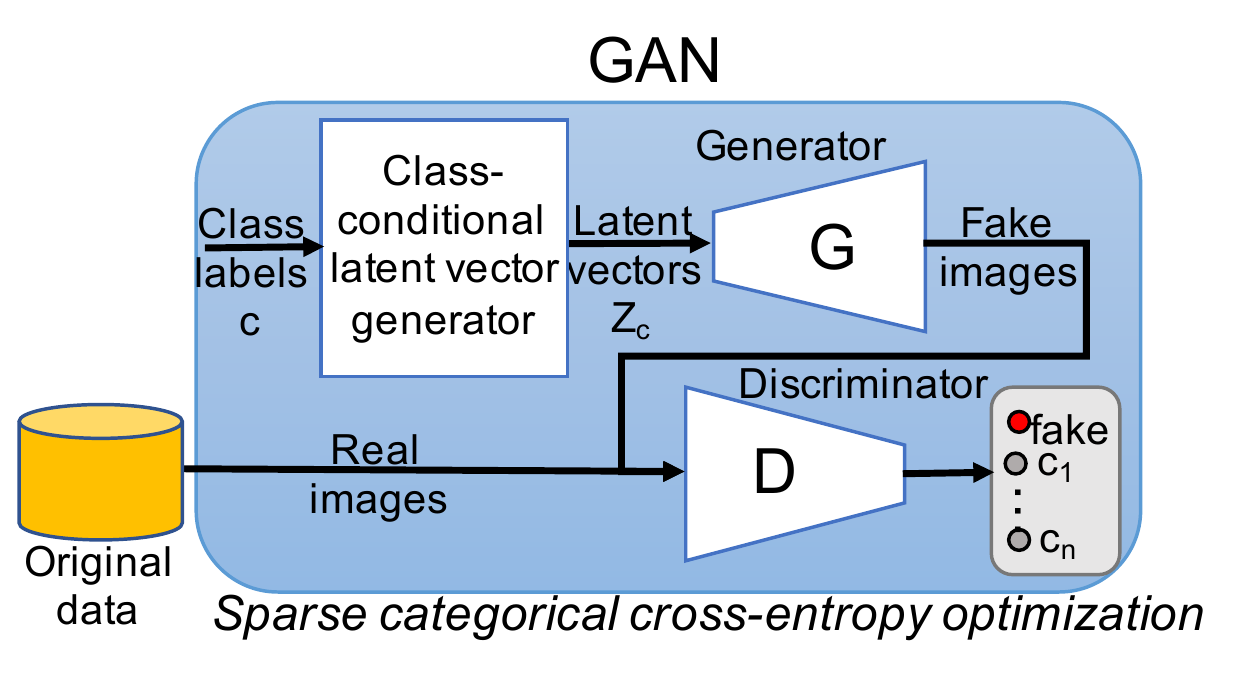}}
            }
\caption{The three training steps of the proposed BAGAN methodology.
}
\label{fig:overview}
\end{figure*}

A different approach is to train the GAN with majority and minority classes jointly 
and to let the GAN distinguish between different classes explicitly.
During training, the generator is explicitly
asked to draw images of every class and to let the discriminator believe that the
generated images are real images of the desired classes. While doing so,
the generator is explicitly rewarded for drawing realistic images
of each class including the minority classes.
To the best of our knowledge, the only method implementing this approach presented in literature so far
is ACGAN \cite{odena2017} where the generator input can be conditioned to draw a target class. In ACGAN,
the discriminator has two outputs,
one to discriminate between \textit{real} and \textit{fake} images $\image$, and the other to classify $\image$ in terms of its class $\class$, Figure \ref{fig:motACGAN}.
During training,
the generator is explicitly asked to draw images $\conditionedFakeImage$ for each class $\class$.
Generator parameters are adjusted to maximize the superposition of two components.
The first component is the log-likelihood of generating an image $\conditionedFakeImage$ considered real by the discriminator.
The second component is the log-likelihood of generating an image $\conditionedFakeImage$ that the discriminator associates with the class $\class$.
We observed that, when a dataset is imbalanced, these two
components become contradictory
for the minority class. This can be explained as follows.
Let us assume that at a point in time the generator converged into a solution where it generates minority-class images with real-word quality. These images would be indistinguishable by the discriminator from the ones in the training dataset.
Since in the training dataset minority-class images are scarce, when a minority-class image is passed to the discriminator during training it is most likely a fake image. To optimize its loss function the discriminator has to associate the \textit{fake} label to all minority-class images.
At this point, the two generator objectives are contradictory and the generator can either draw an image that looks real or one that is representative of the minority class but cannot achieve these two goals at once.
In turn, the generator can be rewarded for drawing images that look real and are not representative of the target minority class. This fact deteriorates the quality of generated images.
Images for the imbalanced MNIST example generated by ACGAN for the $\minorityDigit$ digit are shown in Figure \ref{fig:motResACGAN}.
In this work we propose BAGAN 
that applies class conditioning as ACGAN but differs on the following points.

First, the BAGAN discriminator
has a single output that returns either a problem-specific class label
$\class$ or the label
\textit{fake}, Figure \ref{fig:motBAGAN}.
The discriminator $\discriminator$ is trained for associating to the
images generated by $\generator$ the label \textit{fake}, and to real images $\conditionedImage$
their class label $\class$.
The generator is trained to avoid the \textit{fake} label
and match the desired class labels.
Since this is now defined as a single objective rather than as a superposition of two objectives, by construction it cannot contradict itself and the generator is never rewarded for the generation
of images $\conditionedFakeImage$ that look real if the discriminator does not match them with the desired class label $\class$.

Second,
BAGAN couples GAN and autoencoding techniques to provide a precise selection of the class conditioning and to better avoid mode collapse.
Images for the imbalanced MNIST example generated by BAGAN are of superior quality, Figure \ref{fig:motResBAGAN}.




\section{BAGAN}
\label{sec:met}


The proposed BAGAN methodology aims to generate realistic minority-class images for an imbalanced dataset.
It exploits all available information of the specific classification problem by including in the BAGAN training majority and minority classes jointly.
GAN and autoencoding techniques are  coupled to leverage the strengths
of the two approaches.
GANs generate high-quality images whereas autoencoders converge towards good solution easily \cite{lucic2017}.
Several authors suggest to couple
GANs and autoencoders \cite{srivastava2017,nguyen2016a}.
Nonetheless these works are not directly meant to drive the GAN generative process towards specific
classes. It is not easy to generalize them to enable the GAN
to distinguish between different classes. As explained in the motivating
example, in this work we apply class conditioning as suggested by Odena et al. \cite{odena2017} to embed class knowledge in BAGAN.

We apply a pragmatic use of autoencoders to initialize the GAN
close to a good solution and far from mode collapse. Additionally, we apply the encoder
part of the autoencoder to infer the distribution of the different classes
in the latent space.
The autoencoder-based GAN initialization is achievable by using the same
network topology in the autoencoder and GAN modules, Figures \ref{fig:metAutoencoder} and \ref{fig:metInit}.
The decoding stage $\decoder$ of the autoencoder matches the topology of the
generator $\generator$.
The encoding stage $\encoder$ of the autoencoder
matches the topology of the first layers of the discriminator $\discriminatorEncode$. In BAGAN,
the knowledge in the autoencoder is transferred into the GAN modules
by initializing the parameter weights accordingly, Figure \ref{fig:metInit}.
To complete the discriminator,
a final dense layer $\discriminatorDiscriminate$ with a softmax
activation function translates the latent features into the probability that the image
is \textit{fake} or that it belongs to one of the problem classes
$\class_1$ --- $\class_n$. When the GAN modules are initialized,
a \textit{class-conditional latent vector generator} is set up by learning the probability
distribution of the images in the latent space for the different classes.
Then, all the weights in the generator and discriminator are fine tuned by carrying out a traditional adversarial training, Figure \ref{fig:metGAN}.
Overall, the BAGAN training approach is organized in the three steps 
show in Figure \ref{fig:overview}: \textit{a)} autoencoder training, \textit{b)} GAN initialization,
and \textit{c)} adversarial training.




\textbf{Autoencoder training.} The autoencoder is trained by using all the images in the
training dataset. The autoencoder has no explicit class knowledge, it processes all
images from majority and minority classes unconditionally.
In this work we apply $l_2$ loss minimization for the autoencoder training.

\textbf{GAN initialization.}
Differently from the autoencoder,
the generator $\generator$ and the discriminator $\discriminator$ have
explicit class knowledge.
During the adversarial training, $\generator$ 
is asked to generate images for different classes, and $\discriminator$ is
asked to label the images either as \textit{fake}
or with a problem-specific class label $\class$.
At the moment the GAN is initialized, the autoencoder knowledge is transferred into the GAN modules
by initializing $\generator$
with the weights in the decoder $\decoder$,
and the first layers of the discriminator $\discriminatorEncode$ with the weights of the
encoder $\encoder$, Figure \ref{fig:metInit}.
The last layer of the discriminator $\discriminatorDiscriminate$ is a dense layer with a softmax activation function and generates the final discriminator output.
The weights of this last layer are initialized at random and learnt during the adversarial training.

The discriminator initialization is used
simply to include in $\discriminator$
meaningful features that can help in classifying images.
The initialization of the generator has a deeper reason. When adversarial training
starts, the generator $\generator$ is equivalent to the decoder $\decoder$.
Thus a latent vector $\latentVector$ input to the generator $\generator$ is equivalent to a point in the latent space of the autoencoder, i.e. $\latentVector$ can be seen as the output of $\encoder$ or the input of $\decoder$.
Thus, the encoder $\encoder$ maps
real images into the latent space in use by $\generator$.
We leverage this fact
to learn a good class conditioning before to start the adversarial training,
i.e. we define how a latent vector $\conditionedLatentVector$
should look like for an image of class $\class$.

We model a class in the latent space with a
multivariate normal distribution
$\conditionalLatentDistribution=\mathcal{N}(\meanSymbol_{\class}, \covarianceSymbol_{\class})$ with mean vector
$\meanSymbol_{\class}$ and covariance matrix $\covarianceSymbol_{\class}$.
For each class $\class$, we compute $\meanSymbol_{\class}$ and
$\covarianceSymbol_{\class}$ to match
the distribution of $\conditionedLatentVector=\encoder(\conditionedImage)$ considering
all real images $\conditionedImage$ of class $\class$ available in the training dataset.
We initialize with these probability distributions the \textit{class-conditional
latent vector generator}, that is a random process that takes as input a class label $\class$ and
returns as output a latent vector $\conditionedLatentVector$
drawn at random from $\conditionalLatentDistribution$.
During the adversarial training, the probability distributions $\conditionalLatentDistribution$
are considered invariant forcing the generator
not to diverge from the initial class encoding in the latent space.

\begin{table*}[t!]
\centering
\scriptsize
\caption{Target datasets' information including resolution,
number of classes, and per-class image distribution statistics for the training set.}
\begin{tabular}{|c||c|c||c|c|c|c| }
  \hline
   & & & \multicolumn{4}{c|}{\textbf{Training images per class}} \\ 
   \cline{4-7}
  \textbf{Dataset name} & \textbf{Resolution} & \textbf{Classes} & \textbf{Min} & \textbf{Median} & \textbf{Mean} & \textbf{Max} \\ 
  \hline
  \hline
  \mnist & 28$\times$28 & 10 & 5421 & 5936 & 6000 & 6742 \\ 
  \hline
  \cifar & 32$\times$32 & 10 & 5000 & 5000 & 5000 & 5000 \\ 
  \hline
  \flowers & 224$\times$224 & 5 & 533 & 599 & 634 & 798 \\ 
  \hline
  \gts & 64$\times$64 & 43 & 210 & 600 & 911 & 2250 \\ 
  \hline
\end{tabular}
\label{tab:datasets}
\end{table*}

\begin{figure*}[b!]
	\centering
	\subfigure[Real image samples]{
		\resizebox{\sfigmedium}{!}{\includegraphics{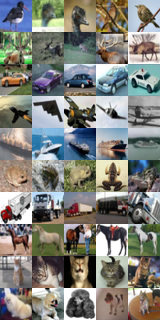}}
	}
	\subfigure[\bagan]{
		\resizebox{\sfigmedium}{!}{\includegraphics{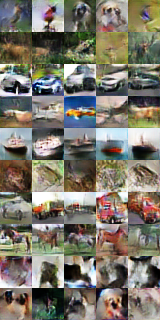}}
	}
	\subfigure[\acgan]{
		\resizebox{\sfigmedium}{!}{\includegraphics{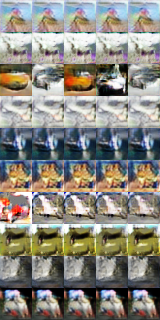}}
	}
	\subfigure[\textit{Simple \gan}]{\label{fig:resRepresentativeCifarGAN}
		\resizebox{\sfigmedium}{!}{\includegraphics{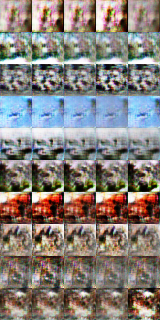}}
	}
	\caption{
		Five representative samples for each class (row) in the \cifar~dataset. For each class, these samples are obtained with generative models trained after dropping from the training set $40\%$ of the images of that specific class.
	}
	\label{fig:resRepresentativeCifar}
\end{figure*}

\textbf{Adversarial training.} During the adversarial training, data flows in batches through
the generator $\generator$ and the discriminator $\discriminator$ and
their weights are fine tuned to optimize their loss functions.
The discriminator classifies an input image as belonging to one of the $n$ problem-specific classes or as being \textit{fake}.
For each batch we supply,
$1/(n+1)$ of the total images are fake, i.e.
we provide the best possible balance for the \textit{fake} class.
The fake data is generated as output of $\generator$ that takes as inputs
latent vectors $\conditionedLatentVector$ extracted from
the {class-conditional
latent vector generator}. In turn, the class-conditional latent vector generator takes as input uniformly distributed
class labels $\class$, i.e. the fake images are uniformly distributed
between the problem-specific classes.
When training the discriminator $\discriminator$ we optimize the
sparse categorical cross-entropy loss function to match the class labels
for real images and the \textit{fake} label for the generated ones. 

For every batch learnt by the discriminator, a batch of the same size is learnt by
the generator $\generator$. To this end, a batch of conditional latent vectors
$\conditionedLatentVector$ is drawn at random by applying a uniform distribution on
the labels $\class$. These vectors are processed by the generator and the output images are fed into the discriminator. The parameters in $\generator$ are optimized
to match the labels selected
by the discriminator with the labels $\class$ used to generate the images.


\section{Results}
\label{sec:res}

We validate the proposed methodology on a set of four datasets.
We consider: \mnist~\cite{mnist}, \cifar~\cite{cifar},
\flowers~\cite{flowers}, and \gts~\cite{gts}.
The former two datasets are well known, \flowers~is a small dataset
including real photos of five categories of flowers
that we reshaped to the resolution of 224x224, and \gts~is
a traffic sign recognition dataset. Details on these datasets are shown in
Table \ref{tab:datasets}. The first three datasets are balanced, \gts~is imbalanced.
We force imbalance in the first three datasets by selecting a class and dropping
a significant amount of its instances from the training set.
We repeat this process for each class and train different generative
models for each resulting imbalanced dataset. The following results for each class
are always obtained when training with that class as minority class and we
refer to the images left out from the training set as \textit{dropped images}.
Since \gts~is already imbalanced, we do not further imbalance it.

\begin{figure*}[t!]
	\centering
	\subfigure[Real image samples]{
		\resizebox{\sfigmedium}{!}{\includegraphics{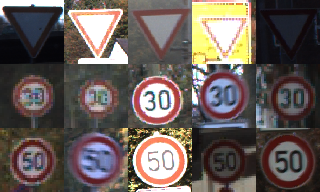}}
	}
	\subfigure[\bagan]{\label{fig:resRepresentativeGtsMajBAGAN}
		\resizebox{\sfigmedium}{!}{\includegraphics{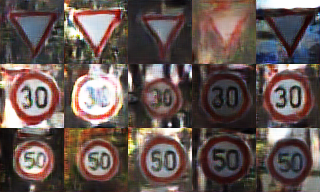}}
	}
	\subfigure[\acgan]{\label{fig:resRepresentativeGtsMajACGAN}
		\resizebox{\sfigmedium}{!}{\includegraphics{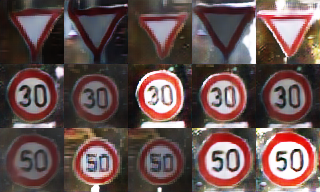}}
	}
	\subfigure[\textit{Simple \gan}]{
		\resizebox{\sfigmedium}{!}{\includegraphics{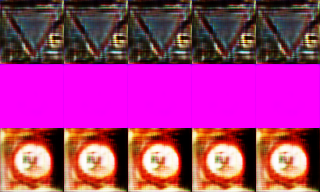}}
	}
	\caption{
		Five representative samples generated for the three most represented majority classes in the \gts~dataset.
	}
	\label{fig:resRepresentativeGtsMaj}
\end{figure*}

\begin{figure*}[t!]
	\centering
	\subfigure[Real image samples]{
		\resizebox{\sfigmedium}{!}{\includegraphics{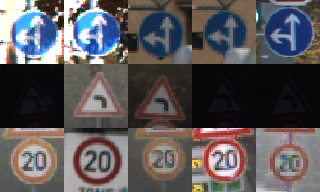}}
	}
	\subfigure[\bagan]{
		\resizebox{\sfigmedium}{!}{\includegraphics{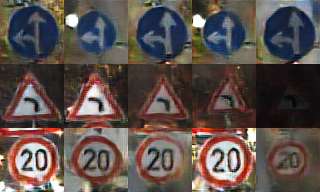}}
	}
	\subfigure[\acgan]{\label{fig:resRepresentativeGtsMinACGAN}
		\resizebox{\sfigmedium}{!}{\includegraphics{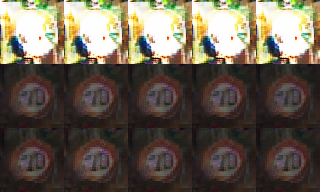}}
	}
	\subfigure[\textit{Simple \gan}]{
		\resizebox{\sfigmedium}{!}{\includegraphics{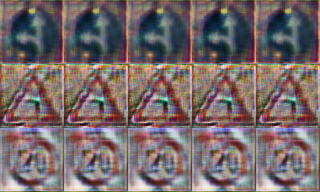}}
	}
	\caption{
		Five representative samples generated for the three least represented minority classes in the \gts~dataset.
	}
	\label{fig:resRepresentativeGtsMin}
\end{figure*}

We compare the proposed \bagan~model with the state-of-the-art \acgan~model \cite{odena2017}. 
To the best of our knowledge, \acgan~is the only
methodology presented in literature so far
to consider class conditioning to draw images of a target class
starting from a dataset including multiple classes (Section \ref{sec:mot}).
Both \bagan~and \acgan~are trained on the target datasets by using
majority and minority classes jointly.
We also consider a simple \gan~approach that
learns to draw the minority-class images by training only on that class.
For a fair comparison, we limit the architecture changes between the considered methodologies (\bagan, \acgan, and \gan). The difference between \bagan~and \acgan~are those described in this paper
(i.e. discriminator output topology and autoencoder-based initialization).
For the simple \gan, we adjust the reference \acgan~discriminator output to discriminate only between real and fake images, and we remove the class conditioning for the generator input (this \gan~is trained only over images from the minority class).
Figures \ref{fig:resRepresentativeCifar}, \ref{fig:resRepresentativeGtsMaj}, and \ref{fig:resRepresentativeGtsMin} show a qualitative analysis of representative images
generated for \cifar~and for the three most and least represented
classes in \gts.
For \cifar~we show results only for minority-class images. For each class, 40\% of that class images are dropped, generative models are trained, and randomly generated images are shown, Figure \ref{fig:resRepresentativeCifar}.
For \cifar, the simple \gan~collapses towards the generation of a single image example per class.
To train this \gan~we
use only 3000 minority-class images (40\% of the minority-class images are dropped and majority classes are not included in the training).
Adversarial networks need many examples to learn drawing new images \cite{gurumurthy2017} and in this case the simple \gan~collapses.
For \acgan~and \bagan~this issue is less relevant because they can learn features from minority and majority classes jointly.
To better understand the different behavior of \acgan~and \bagan, let us focus on the \gts~dataset Figures \ref{fig:resRepresentativeGtsMaj} and \ref{fig:resRepresentativeGtsMin}. This dataset
is originally imbalanced and we train the generative models without modifying it.
For the majority classes, both
\acgan~and \bagan~return high-quality results,
Figures \ref{fig:resRepresentativeGtsMajACGAN} and \ref{fig:resRepresentativeGtsMajBAGAN}. Nonetheless, \acgan~fails in drawing
images for the minority classes and collapses towards the generation of a
single example for each class, Figure \ref{fig:resRepresentativeGtsMinACGAN}.
In some cases \acgan~produces images that are not
representative of the desired class, e.g.
the second row in Figure \ref{fig:resRepresentativeGtsMinACGAN}
should be a warning sign whereas a speed limit is drawn.
\bagan~is never rewarded for drawing a realistic image if this does not represent
the desired class. Thus, \bagan~does not exhibit this behavior.

\begin{figure*}[b!]
\centering
            \resizebox{\sfigwhole}{!}{\includegraphics{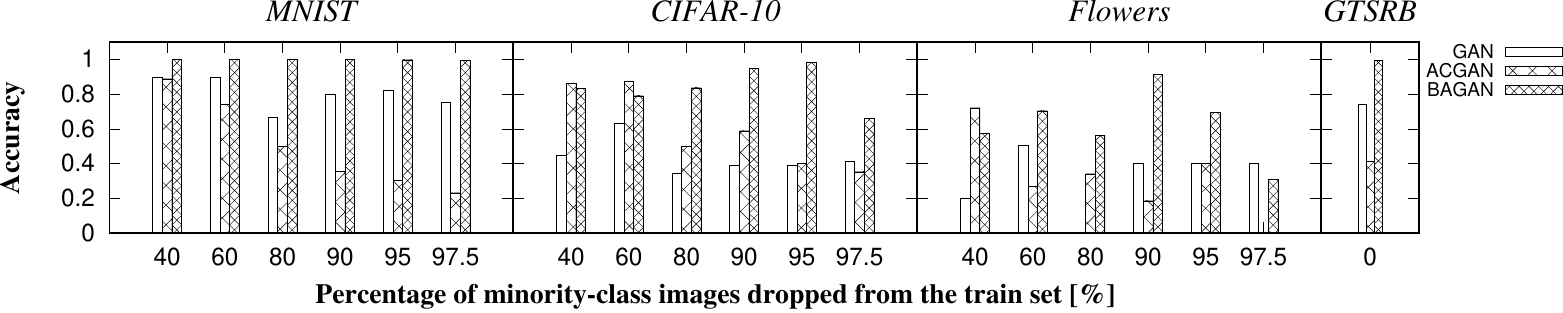}}
\caption{Accuracy of the images generated by the considered methodologies when varying the percentage of minority-class images dropped before training the generative models. The accuracy is based on a \resnet~classifier trained without dropping any image.
}
\label{fig:resAugAccuracy}
\end{figure*}



\subsection{Quantitative Assessment of Generated Images}

Since our goal is to leverage the generative model to augment an imbalanced dataset
by generating additional minority-class images,
we aim at the following goals:
\begin{enumerate}[a)]
\item Generated images must represent the desired class. \label{res:accurate}
\item Generated images must not be repetitive. \label{res:noCollapse}
\item Generated images must be different from the real ones already available in the training set. \label{res:noOverfit}
\end{enumerate}

Missing to meet \textit{\ref{res:accurate})} means
that the generative model is not capable to generate images that accurately represent
the target class and they look either as real examples of other classes or they do not
look real.
Missing to meet \textit{\ref{res:noCollapse})} means that the generative model collapsed
to the generation of a single or few modes. 
Missing to meet \textit{\ref{res:noOverfit})}
means that we simply learnt to redraw the available training images.
We assess the quality of the generated images on the basis of these three goals.

\begin{figure*}[t!]
\centering
            \resizebox{\sfigwhole}{!}{\includegraphics{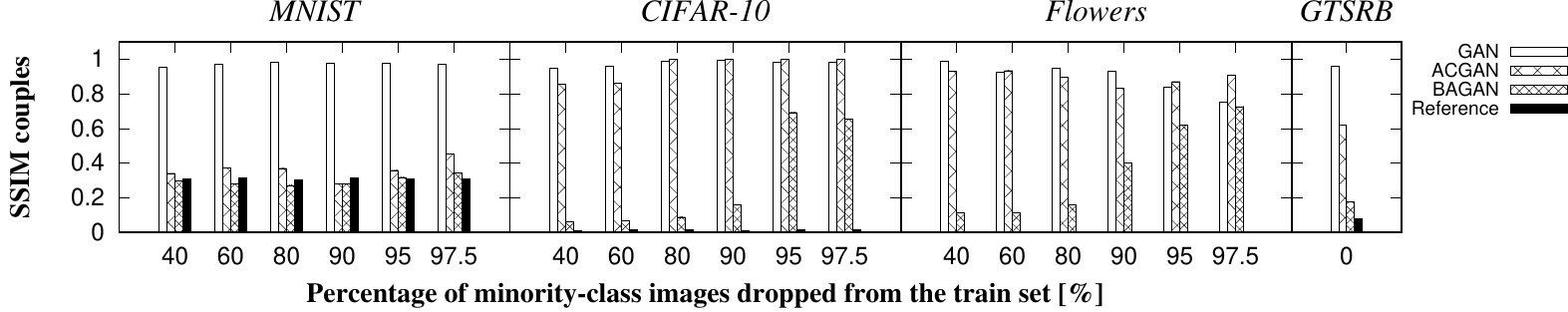}}
\caption{Structural similarity for generated image couples (SSIM couples, $y$ axis) when varying the percentage of images dropped from the training set ($x$ axis).
}
\label{fig:resNoCollapse}
\end{figure*}

\begin{figure*}[b!]
\centering
            \resizebox{\sfigwhole}{!}{\includegraphics{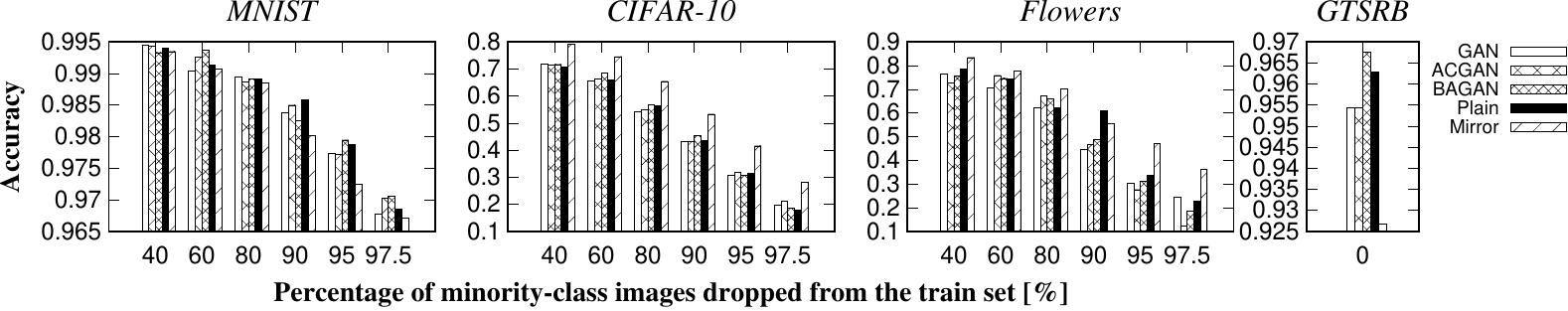}}
\caption{Average accuracy for the minority-class achieved with a \resnet~classifier trained with the augmented dataset whose balance is restored after dropping a percentage of minority-class images.
}
\label{fig:resFinal}
\end{figure*}

\textbf{Accuracy of the generated images.} To verify that the images generated by
the considered methodologies are representative of the desired classes,
we classify them by means of a deep learning model trained on the whole original dataset and we verify if the predicted classes match
the target ones. In this work we use a \resnet~model \cite{resnet}. Results are shown in Figure \ref{fig:resAugAccuracy}. The simple \gan~returns the worst accuracy for generated images.
The proposed \bagan~approach is generally better than the other approaches and generates images that the \resnet~model can classify with the highest accuracy.
We observe again that a strong imbalance can significantly deteriorate the quality of generated images with an accuracy that decreases as the percentage of dropped images increases. This phenomenon is most evident for
 \acgan~when targeting the \mnist~dataset.

\textbf{Variability of generated images.}
We measure similarity between two images by means of the
structural image similarity \ssim~\cite{ssim}.
This metric predicts human perceptual similarity judgment, it returns one when the two images are identical
and decreases as differences become more relevant.
To verify that
generated images are diverse, for each class we repeatedly generate a couple of images and measure their similarity \ssim. 
Figure \ref{fig:resNoCollapse} shows this diversity analysis for the considered datasets averaged over all classes. For \mnist, \cifar, and \flowers, we vary the percentage of minority-class images dropped within the set $\{40, 60, 80, 90, 95, 97.5\}$, whereas for \gts~we use the originally imbalanced dataset.
We include in the analysis also a reference value
that is the average \ssim~between real image couples of the same class.
When taking a random couple of real images
for \cifar~or \flowers, these have so little in common that
the reference \ssim~gets very close to zero.
In general real images are always more variable than the generated ones (lower \ssim).
Variability in images generated by the simple \gan~approach is very little and sampled image couples have \ssim~very close to one. The proposed \bagan~methodology
exhibit the best variability with respect to \gan~and \acgan with \ssim~values closest
to the reference. For \cifar~and \flowers, all methodologies deteriorate for strong imbalances with \ssim~values that increase with the percentage of images dropped from the training set.

\textbf{Image diversity with respect to the training set.}
To assess the variability of generated images with respect to the ones already available in the training set.
We compute the \ssim~between
generated images and their closest real neighbour.
We compare this
value with respect to the image variability in the training set, i.e. the \ssim~value between a real image and its closest real neighbour.
These \ssim~values are very close to each others meaning that there was no overfitting. This statement holds for all the considered methodologies. In particular \ssim~values of about 0.8, 0.25, 0.05, and 0.5 are measured respectively for \mnist, \cifar, \flowers, and \gts. 




\subsection{Quality of the Final Classification}

We finally assess the accuracy of a deep-learning classifier trained on an augmented dataset. For \mnist, \cifar, and \flowers, for each class we:
\textit{1)} select this class as minority class, \textit{2)} generate an imbalanced dataset by dropping a percentage of images for this class from the training set, \textit{3)} train the considered generative models, \textit{4)} augment the imbalanced dataset to restore its balance by means of the generative models, \textit{5)} train a \resnet~classifier for the augmented dataset, and \textit{6)} measure the classifier accuracy for the minority class over the test set. Since \gts~is already imbalanced, for this dataset we skip steps \textit{1)} and \textit{2)}.
Augmentations obtained by the generative models are compared to the \textit{plain} imbalanced dataset  and to an horizontal mirroring augmentation  approach (\textit{mirror}) where new minority-class images are generated by mirroring the ones available in the training set.

Accuracy results averaged over the different classes are shown in Figure \ref{fig:resFinal}. The proposed \bagan~methodology returns the best accuracy for \gts~and most of the time also for \mnist. These two datasets are characterized by features sensible to the image orientation and the mirroring approach as expected returns the worst accuracy results because it disrupts these features. For \cifar~and \flowers~the best accuracy is achieved by using the mirroring approach. Mirroring for these datasets does not disrupt any feature, qualitatively the mirrored images are as good as the original ones. The \bagan~approach still provides the best accuracy when compared to \acgan~and \gan.

From this analysis we conclude that \bagan~is superior to other state-of-the-art adversarial generative networks when aiming at the generation of minority-class images starting from an imbalanced dataset. Additionally we conclude that: when it is not easy to augment a dataset with traditional techniques because of orientation-related features, \bagan~can be applied to improve the final classification accuracy.

\nocite{dosovitskiy2016}

\section{Conclusion}

In this work we presented a methodology to restore the balance of an imbalanced dataset by using generative adversarial networks. In the proposed BAGAN framework the generator and the discriminator modules are initialized by means of an autoencoder to start the adversarial training from a good solution and to learn how different classes should be represented in the latent space.

We compared the proposed methodology against the state of the art. Empirical results demonstrate that BAGAN is superior to other generative adversarial networks when aiming at the generation of high quality images starting with an imbalanced training set. This in turn
results in a higher accuracy of deep-learning classifiers trained over the augmented dataset where the balance has been restored.

\bibliography{biblio}
\bibliographystyle{unsrt}

\end{document}